\documentclass[10pt,twocolumn,letterpaper]{article}

\usepackage[pagenumbers]{cvpr} 

\usepackage[dvipsnames]{xcolor}

\usepackage{booktabs}
\usepackage{multirow}
\usepackage{amsmath}
\usepackage{amssymb}
\usepackage{diagbox}
\usepackage[accsupp]{axessibility} 

\usepackage{pifont}
\newcommand{\cmark}{\ding{51}}
\newcommand{\xmark}{\ding{55}}

\newcommand{\ra}[1]{\renewcommand{\arraystretch}{#1}} 

\definecolor{cvprblue}{rgb}{0.21,0.49,0.74}
\usepackage[pagebackref,breaklinks,colorlinks,citecolor=cvprblue]{hyperref}

\def\paperID{010} 
\def\confName{CVPR}
\def\confYear{2024}

\title{NOISe: Nuclei-Aware Osteoclast Instance Segmentation\\for Mouse-to-Human Domain Transfer}

\author{Sai Kumar Reddy Manne$^{1,\star}$ ~ Brendan Martin$^{1,\star}$ ~ Tyler Roy$^2$ ~ Ryan Neilson$^2$ \\
Rebecca Peters$^{2,3}$ ~ Meghana Chillara$^1$ ~ Christine W. Lary$^1$ ~ Katherine J. Motyl$^{2,3,4}$\\
Michael Wan$^{1,\dagger}$\\ 
${^1}$Northeastern University ~ ${^2}$MaineHealth Institute for Research \\
${^3}$University of Maine ~ ${^4}$Tufts University School of Medicine\\
${^\star}$equal contribution ~ ${^\dagger}$\tt \small mi.wan@northeastern.edu}

\begin{document}

\twocolumn[{%
\renewcommand\twocolumn[1][]{#1}%
\maketitle
\begin{center}
    \centering
    \captionsetup{type=figure}
    \includegraphics[width=0.9\linewidth]{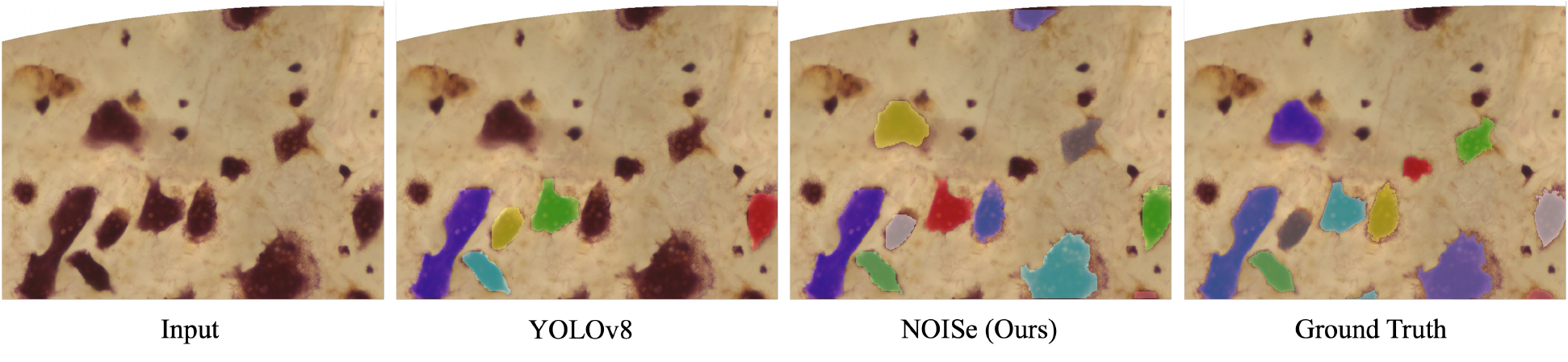}
    \captionof{figure}{Human osteoclast cell segmentations from our nuclei-aware osteoclast instance segmentation (NOISe) model, compared to a baseline YOLOv8 model. Since both models are trained only on mouse osteoclast data, this highlights the effectiveness of our nuclei-aware pretraining strategy for transfer learning to the human domain, where osteoclast microscope samples are harder to obtain and annotate.}
\end{center}%
}]

\begin{abstract}
Osteoclast cell image analysis plays a key role in osteoporosis research, but it typically involves extensive manual image processing and hand annotations by a trained expert. In the last few years, a handful of machine learning approaches for osteoclast image analysis have been developed, but none have addressed the full instance segmentation task required to produce the same output as that of the human expert led process. Furthermore, none of the prior, fully automated algorithms have publicly available code, pretrained models, or annotated datasets, inhibiting reproduction and extension of their work. We present a new dataset with ${\sim}2\times 10^5$ expert annotated mouse osteoclast masks, together with a deep learning instance segmentation method which works for both \textit{in vitro} mouse osteoclast cells on plastic tissue culture plates and human osteoclast cells on bone chips. To our knowledge, this is the first work to automate the full osteoclast instance segmentation task. Our method achieves a performance of 0.82 $\text{mAP}_{0.5}$ (mean average precision at intersection-over-union threshold of $0.5$) in cross validation for mouse osteoclasts. We present a novel \textbf{n}uclei-aware \textbf{o}steoclast \textbf{i}nstance \textbf{se}gmentation training strategy (\textbf{NOISe}) based on the unique biology of osteoclasts, to improve the model's generalizability and boost the $\text{mAP}_{0.5}$ from 0.60 to 0.82 on human osteoclasts. We publish our annotated mouse osteoclast image dataset, instance segmentation models, and code at \texttt{github.com/michaelwwan/noise} to enable reproducibility and to provide a public tool to accelerate osteoporosis research\footnote{We gratefully acknowledge support from the National Institutes of Health under award numbers R01AR076349, R01AR081040, and P20GM121301; the Roux Institute at Northeastern University; and AI + Health, Institute for Experiential AI, Northeastern University.}.

\end{abstract}    
\section{Introduction}
\label{sec:intro}

Osteoporosis is a widespread, debilitating bone structure disease, affecting an estimated 200 million people worldwide, especially older women. Discovery of treatments for osteoporosis hinges around advances in basic and translational research in bone biology, and in particular, in the delicate balance between bone-synthesizing osteoblast cells and bone-resorbing \textit{osteoclast} cells, which is disrupted in osteoporosis \cite{burns_direct_2016,solomon_osteoporosis_2014}. Many drugs, including alendronate and denosumab, target a reduction of bone resorption by modulating the activity of osteoclasts \cite{bone_ten_2004,cummings_denosumab_2009}. The potential effectiveness of candidate treatments are gauged by applying them to osteoclast cell cultures and observing their effects on the differentiation and function of those cells, via cellular microscopy. These effects are quantified by a laborious, low-throughput, manual procedure, in which cell shapes are carefully annotated on microscope images by human researchers, with the help of image processing tools that require extensive finetuning and manual adjustment. In addition, high measurement variability necessitates many replicates, further exacerbating the problem. Even a small translational treatment study can require over 1,000 hours of skilled manual labor to analyze.

\begin{figure}
    \centering
    \includegraphics[width=0.8\linewidth]{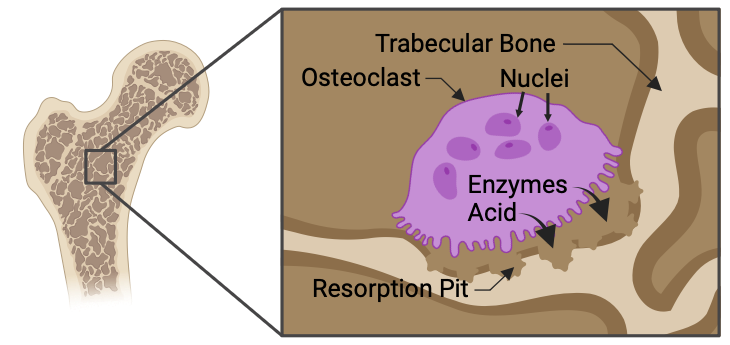}
    \caption{An illustration of an osteoclast cell in purple acting on bone structure in light brown. Osteoclasts are responsible for bone-resorption, and are characterized by having three or more nuclei and being positive for purple TRAP enzyme stain.}
    \label{fig:osteoclast}
\end{figure}

Fortunately, initial strides have been made in automating osteoclast microscope image characterization with machine learning, with some work targeting osteoclast cell \textit{object detection} and some attempting osteoclast \textit{semantic segmentation} \cite{cohen-karlik_quantification_2021,emmanuel_artificial_2021,wang_oc_finder_2022,kohtala_automated_2022,davies_machine_2023}. Critically, though, none of these efforts replicate the full annotation task required by osteoporosis treatment research, namely, that of osteoclast \textit{instance segmentation}, which entails both osteoclast detection and \textit{individual} cell segmentation. Furthermore, prior approaches are variously limited by (i) requiring manual image processing and thereby not being truly automated, (ii) considering the more readily available mouse osteoclast cells and ignoring the human osteoclast cells needed for treatment research, and (iii) requiring specific cellular stains or lab preparations. Lastly, but no less critically, none of these earlier groups addressing object detection or semantic segmentation publish their code, models, or annotated datasets, preventing reproduction and validation of their work, and implementation by osteoporosis researchers. 

To address these barriers, we present the first public osteoclast dataset with instance segmentation masks meticulously annotated for ${\sim}2\times 10^5$ osteoclasts, and a fully-automated algorithm which solves the osteoclast instance segmentation task. Our deep learning method uses a state of the art instance segmentation architecture based on convolutional neural networks, further improved by a nuclei-aware training strategy, injecting domain knowledge from cellular biology into the learning process. We test the model's generalizability by testing on human osteoclast cells imaged on a bone chip and show a performance improvement of $\text{mAP}_{0.5}$ from 0.60 to 0.82. In summary, our key contributions are as follows.
\begin{enumerate}
    \item We publish the first public dataset for osteoclast instance segmentation, consisting of microscope images of mouse osteoclast cultures derived from osteoporosis treatment experiments, fully expert annotated with individual osteoclast locations and shape masks.
    \item We publish our osteoclast instance segmentation models. To our knowledge, these are the first fully-automatic osteoclast detection or segmentation algorithms made publicly available.
    \item We describe a novel \textbf{n}uclei-aware \textbf{o}steoclast \textbf{i}nstance \textbf{se}gmentation strategy (\textbf{NOISe}) aimed at improving the generalizability of our instance segmentation model for mouse osteoclasts on plastic tissue culture plates and human osteoclasts on bone chips.
\end{enumerate}

\section{Background}
\label{sec:background}

\subsection{Bone Biology and Research}

\begin{table*}
\centering
\caption{Prior Algorithmic Approaches to Osteoclast Microscope Image Analysis}
\label{tab:prior_methods}
\resizebox{\textwidth}{!}{
\ra{0.8}
\begin{tabular}{p{1.4in}rp{1.4in}ccccccccc}
\toprule
& & & \textbf{Fully}  & \textbf{Object} & \textbf{Semantic} & \textbf{Instance} & \textbf{Human} &
\textbf{Standard} &
\textbf{Data} & 
\textbf{Code}\\ 
\textbf{Paper} & \textbf{Year} & \textbf{Dataset Size} & \textbf{Automated?} & \textbf{Detection?} & \textbf{Segmentation?} & \textbf{Segmentation?} & \textbf{Cells?} & \textbf{Stains?} & \textbf{Public?} & \textbf{Public?}  \\ \midrule
Emmanuel \textit{et al.} \cite{emmanuel_artificial_2021} & 2021 & Slides from 20 rat tibiae  & \xmark & \cmark & \xmark & \xmark & \xmark & \cmark & \xmark & \xmark \\ \midrule
Cohen--Karlik \textit{et al.} \cite{cohen-karlik_quantification_2021} & 2021 & ${\sim}10^4$ mouse osteoclasts & \cmark  & \cmark & \xmark & \xmark & \xmark & \cmark & \xmark & \xmark \\ \midrule
Wang \textit{et al.} \cite{wang_oc_finder_2022} & 2022 & ${\sim} 1.4\times 10^4$ mouse osteoclasts  & \xmark & \xmark & \cmark & \xmark & \xmark & \cmark & \cmark & \cmark \\ \midrule
Kohtala \textit{et al.} \cite{kohtala_automated_2022} & 2022 & ${\sim} 10^5$ mouse osteoclasts  & \cmark & \cmark & \xmark & \xmark & \cmark & \xmark & \xmark & \xmark \\ \midrule
Davies \textit{et al.} \cite{davies_machine_2023} & 2023 & ${\sim}10^5$ mouse osteoclasts & \xmark & \xmark & \cmark & \xmark & \xmark & \cmark & \xmark & \xmark \\ \midrule
\textbf{Ours}            & 2024 & ${\sim} 2\times 10^5$ mouse,  ${\sim} 4\times 10^4$ human osteoclasts & \cmark & \cmark  & \cmark & \cmark & \cmark & \cmark & \cmark & \cmark                    \\ \bottomrule
\end{tabular}
}
\end{table*}

Healthy, homeostatic bone remodeling is mediated by a delicate balance between osteoblasts, which are derived from mesenchymal stem cells and responsible for bone formation, and \textit{osteoclasts}, which are multi-nucleated cells derived from the macrophage and monocyte lineages, and are responsible for bone resorption \cite{burns_direct_2016,solomon_osteoporosis_2014}. Osteoclasts are dynamical cells which, through their lifetime, can undergo fusion, fission, recycling, and apotosis \cite{mcdonald_osteoclasts_2021}. Osteoclast dynamics and function are disrupted in diseases of the bone such as rheumatoid arthritis and osteoporosis, spurring intense research interest in the life cycle and morphology of osteoclasts. Osteoclasts can be studied via microscopy: tartrate-resistant acid phosphatase (TRAP) enzyme stain highlights certain cells in purple, and these are classified as osteoclasts if they have three or more nuclei, and pre-osteoclasts if they have less than three nuclei. See \cref{fig:osteoclast} for an illustration of an osteoclast cell under TRAP staining.

Traditionally, bone and osteoporosis researchers have carried out osteoclast microscope image analysis with a combination of manual annotation and manually-developed image processing pipelines, using specialized software for biological image analysis such as ImageJ or CellProfiler \cite{mcdonald_osteoclasts_2021,brent_contemporary_2022}. This process requires training and experimentation, and manual annotations are unavoidable. Each cell culture well takes 2--3 hours of labor to process, after training. We typically produce at least 3 biological replicates for each manipulation (\textit{e.g.}, genotype or drug treatment), themselves with 5 technical replicates each to account for variability, resulting in at least 15 wells per experimental group. As a result, even a small translational treatment study can require over 1,000 hours of labor to process.

\subsection{Osteoclast Image Analysis Algorithms}

The need for an automated solution to osteoclast microscope image analysis is reflected by the flurry of recent research activity applying machine learning tools to this task, summarized in \cref{tab:prior_methods}. However, while some methods tackle osteoclast detection or semantic segmentation, none of them address the full instance segmentation task that would provide the same detailed output that the manual process does, namely, the individual locations, shapes, and areas of each osteoclast cell. Secondly, and equally critically, no other group has published model code or software that would allow other researchers to use their osteoclast detection or semantic segmentation methods, nor annotated datasets that would allow others to reproduce their results or benchmark detection or segmentation methods. 
We will address both of these critical shortcomings in our work.

The first machine learning paper for osteoclast analysis, by Emmanuel \textit{et al.} \cite{emmanuel_artificial_2021}, provides a step-by-step guide on how to use a piece of commercial machine learning software to segment osteoclasts \textit{in vivo} from slides featuring thin rat tibiae bone slices. However, many steps involve manual choices in response to intermediate results, so the method is far from automated. Furthermore, the nature of the dataset used and the corresponding train--test split are not specified, and validation is only provided by comparing software and human estimations of a single quantity derived from the segmentations over the entire dataset, so verification of the method’s performance on osteoclast identification is not possible.

Cohen-Karlik \textit{et al.} \cite{cohen-karlik_quantification_2021} take a more standard approach, studying \textit{in vitro} slides featuring primary osteoclasts differentiated from bone marrow cells extracted from female mice femurs and tibias and stained with TRAP. They train a convolutional neural network (VGG16) on osteoclast object detection using image patches extracted from 10 culture wells, bolstered by extensive data augmentation, and test on one held-out well, reporting high correlation scores between human and machine-predicted osteoclast counts and areas. However, high correlation does not rule out systematic errors in the predictions---for instance, they use the area of the estimated box bounding the osteoclast as a proxy for the cell's area, which even in ideal conditions will yield a consistent overestimate of the true area (\textit{e.g.,} if osteoclasts were circular, by a factor of $\frac{4}{\pi}\cong1.3$). Standard machine learning metrics such as mean average precision ($\text{mAP}_t$) could help detect and quantify such prediction biases, but the authors do not publish their model or dataset, preventing independent evaluation.

Wang \textit{et al.} \cite{wang_oc_finder_2022} 
develop a two-step method: they first apply a non-learning-based image processing and size filtration step to obtain a set of candidate cell locations and shapes, and then second apply a machine learning classifier to determine which of the candidate cells are osteoclasts. Since the first step requires manual tuning, the algorithm is not fully automated. Furthermore, while the authors do publish part of their dataset and algorithm for public use, the data only supports training and verification of the second step. Specifically, they only provide the coordinates of the approximate centers of osteoclast cells, and only those successfully detected in the first image processing step. Thus, their method does not automate osteoclast detection or instance segmentation, and likewise, their data does not directly support training or testing of osteoclast detection or instance segmentation methods.


\begin{figure*}
    \centering
    \includegraphics[width=1\linewidth]{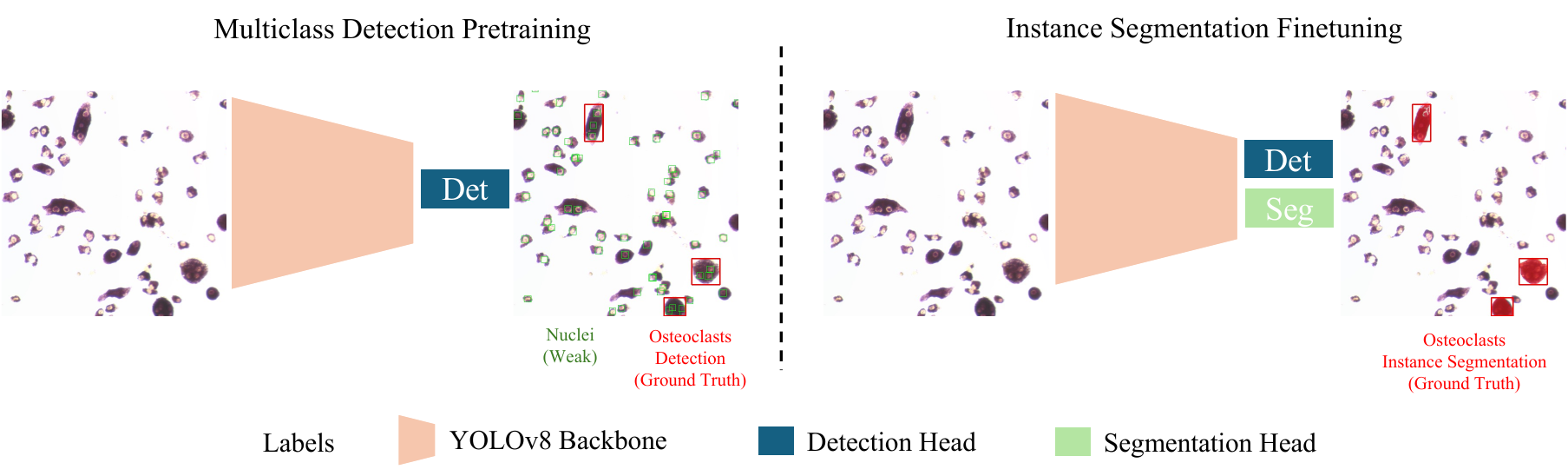}
    \caption{An overview of NOISe, our nuclei-aware osteoclast instance segmentation training pipeline. A two-stage training process features a pretraining stage for multiclass detection weakly supervised by nuclei location information. The pretraining significantly boosts subsequent performance of the overall osteoclast instance segmentation model, especially in the data-scarce human domain.}
    \label{fig:method}
\end{figure*}
Kohtala \textit{et al.} \cite{kohtala_automated_2022}
published the first method algorithmic method specifically designed for human osteoclasts. They use 307 wells differentiated from both CD14+ and macrophage cells, which are split into various train--test configurations, and train a YOLOv4-based DarkNet model optimized for object detection. Detection results are reported using $\text{mAP}_t$ and performance reaches $\text{mAP}_{0.1} = 0.85$ and $\text{mAP}_{0.5} = 0.76$; overall counts show high concordance with human annotations. This work offers an approach in line with standard machine learning practices, yielding interpretable, robust, and comparable metrics and reproducible methods. However, unique among algorithmic approaches, this method uses cell samples obtained from a fluorescent nuclear staining process, which is intended aid manual image analysis and therefore likely makes the object detection task easier. As such, there is no reason to assume that the method will work with cell samples obtained using the standard TRAP stain, but we cannot verify this since the code and data are not made public.

The most recent method, by Davies \textit{et al.} \cite{davies_machine_2023}, is largely interested in analyzing changes to osteoclasts in response to treatments, using results obtained from standardized algorithms. The latter consists of an open-sourced deep learning image segmentation tool, ilastik, followed by semi-manual processing with another open-sourced tool, ImageJ, to obtain desired osteoclast characteristics. Their evaluation metrics are fairly unique, in this case relying on correlation scores on aggregate metrics like total osteoclast cell count and area. This again precludes reproducibility and robust comparison with other methods. Data and models are not provided publicly. 

\section{Method}
\label{sec:method}

\subsection{End-to-End Instance Segmentation}
\label{sec:instance-segmentation}

Instance segmentation identifies unique instances of objects in an image and predicts their segmentation masks. It can be thought of as the combination of object detection, which locates and distinguishes instances of objects of interest with minimal bounding boxes, and semantic segmentation, which classifies each pixel or region by object type (or as belonging to the background). Sometimes, the methods for accomplishing both tasks can overlap, while in other contexts, the two can be addressed separately, as is often the case in cytological and histological imaging. As detailed in \cref{sec:background}, all previous approaches to algorithmic osteoclast image analysis have attempted either object detection, or semantic segmentation, or their own unique task, but none have directly tackled instance segmentation.

The core of our osteoclast instance segmentation is the YOLOv8 model \cite{Jocher_Ultralytics_YOLO_2023}, which follows a long line of You Only Look Once (YOLO) object detection models \cite{redmon2016you}. YOLO models are based on the idea of directly regressing bounding boxes in an end-to-end manner, supplanting earlier multi-stage methods \cite{girshick2015fast}. YOLOv8 predicts a segmentation mask, a bounding box, a classification label, and an ``objectness'' score for each instance. The model architecture employs a version of the CSPDarknet53 backbone first introduced in YOLOv4 \cite{bochkovskiy2020yolov4} and output heads for bounding box regression, class prediction, and two heads for segmentation \cite{Terven_2023}. The segmentation heads are similar to the You Only Look At Coefficients (YOLACT) model \cite{yolact++}: one head predicts 32 mask prototypes, and the other head predicts 32 mask coefficients. The final instance segmentation predictions come from a linear combination of the prototypes weighted by the coefficients, followed by score thresholding and non-maximum suppression (NMS).

\begin{figure*}
    \centering
    \includegraphics[width=0.8\linewidth]{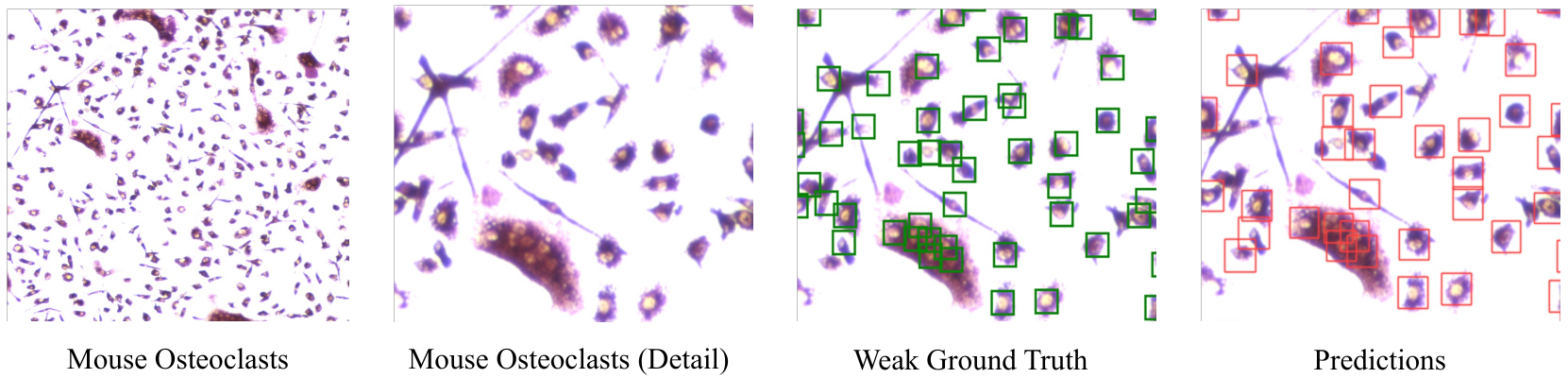}
    \vspace{-2mm}
    \caption{Detail from a mouse osteoclast microscope image, together with weak ground truth nuclei labels (note the uniform size and shape of the boxes), and YOLOv8 object detection predictions of the same. Our nuclei-aware training method exploits this weak ground truth information to improve the generalizability of osteoclast instance segmentation.}
\label{fig:nuclei_detection_examples}
\end{figure*}

YOLOv8 is trained with a combination of four loss functions: (i) a classification loss $L_\text{cls}$, which is a simple binary cross-entropy (BCE) calculated separately for each class; (ii) a complete intersection over union (CIoU) \cite{Zheng_Wang_Liu_Li_Ye_Ren_2020} loss $L_\text{box}$, which compares the central points and aspect ratios of the two boxes along with IoU for the bounding box regression; (iii) a distribution focal loss (DFL) \cite{dist_focal_loss} $L_\text{dfl}$, which distinguishes itself from classic bounding box regression by targeting a distribution of correct bounding boxes as opposed to a single correct target; and (iv) a segmentation loss $L_\text{seg}$, which is a pixel-wise BCE similar to that of YOLACT. The final loss function is a weighted combination of the four components,
\begin{equation}
    L = \lambda_\text{box} L_\text{box} + \lambda_\text{cls} L_\text{cls} + \lambda_\text{dfl} L_\text{dfl} + \lambda_\text{seg} L_\text{seg},
\end{equation}
with corresponding weights $\lambda_\star$ for each component $L_\star$.

\subsection{Nuclei-Aware Osteoclast Instance Segmentation (NOISe)}

Most osteoclast image analysis algorithms address the domain of \textit{in vitro} mouse (or generally murine) osteoclast cells in clear plastic plates, due to availability and controllability of mouse subjects for research. However, translational osteoporosis researchers must also analyze the less commonly available human osteoclast cells, which in the case of our dataset are hosted and imaged on thin transluscent bone chips to study the effects of treatments on osteoclast bone resoprtion. In our experimentalists' experience, human osteoclasts have more complex shapes and take longer to annotate manually. Since we have enough mouse osteoclast data to support strong mouse osteoclast instance segmentation, we explore the possibility of improving human domain performance using only a small amount of human osteoclast data for training and testing. The one prior machine learning osteoclast image analysis work that considers human osteoclasts (on transparent plates) \cite{kohtala_automated_2022} uses fluorescent Hoechst staining to highlight the nuclei. Given the importance of nuclei in the definition and identification of osteoclast cells, this fluorescent staining presumably makes osteoclast detection easier, and at the same time, less applicable to the more common plain light setting. Here, we describe a machine learning alternative to fluorescent staining for highlighting nuclei.

Our \textbf{n}uclei-aware \textbf{o}steoclast \textbf{i}nstance \textbf{se}gmentation (NOISe) pretraining--finetuning strategy, shown in \cref{fig:method}, leverages weak or pseudo nuclei information for efficient transfer learning. First, we curate a small, informal nuclei detection dataset as described in \cref{sec:nuclei_dataset}. We train a YOLOv8 object detection model on the curated nuclei dataset, which achieves an $\text{mAP}_{0.5}$ of 0.85, and demonstrates highly precise nuclei detection, as shown in \cref{fig:nuclei_detection_examples}. We next apply this detector to our entire mouse osteoclast dataset, providing a set of weak ground truth nuclei bounding box labels, on top of the existing ground truth osteoclast instance segmentation labels. Then, in order to inject the nuclei information in our osteoclast segmentation model, we pretrain it on a multiclass object \textit{detection} task with both osteoclast labels and weak nuclei labels. This encourages the model to learn priors about nuclei and their relationship with osteoclasts. Finally, we finetune the pretrained detection model for instance segmentation task with osteoclast segmentation labels. Since the nuclei bounding boxes can only be considered as weak supervision signals, their use must be justified by improvements on downstream tasks that are validated with accurate ground truth instance masks. We show experimentally, in \cref{sec:human-results}, that the use of the nuclei in the NOISe method does indeed lead to segmentation improvements over a baseline YOLOv8 model.

\section{Datasets}
\label{sec:datasets}

\begin{figure*}
    \centering
    \includegraphics[width=0.95\linewidth]{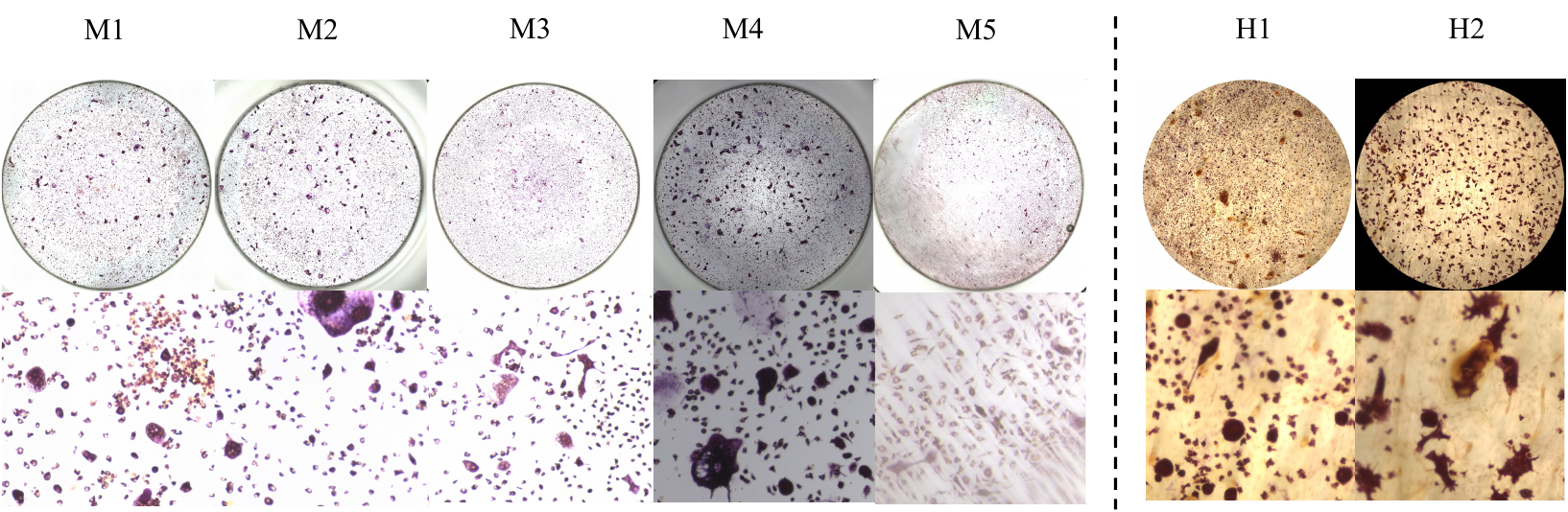}
    \vspace{-2mm}
    \caption{Cell images and cropped patches from mouse experiments M1 through M5, and human experiments H1 and H2, illustrating the diversity in slide lighting, background, and appearance, and in osteoclast size, shape, and density in our data.}
    \label{fig:dataset}
\end{figure*}

\subsection{Expert-Annotated Human and Mouse Osteoclast Instance Segmentation Dataset}
Microscope images with osteoclast cells were captured from two types of experiments, those with mouse osteoclasts plated on 96-well plastic plates to measure differentiation of osteoclasts in response to treatments, and those with human osteoclasts plated on bone chips (devitalized bovine bone slices within a 96-well plastic plate) to measure bone resorption and differentiation in a more realistic setting. After one week of differentiation for mouse cells, and two weeks for human cells, wells were fixed and treated with tartrate-resistant acid phosphatase (TRAP). Treated and control plates were photographed under the microscope at approximately $10^5\times 10^5$ resolution (10 billion pixels, or 10 gigapixels). Individual osteoclast cell instance masks were annotated by human researchers with at least two years of experience in osteoclast culturing, imaging, and osteoclast image analysis, after passing an osteoclast cell count validation test. Using ImageJ, a purple color filter was applied and selected objects with areas of at least 2,000 pixels (the rough mean size of a 3-nuclei osteoclast) were marked as osteoclast candidates. Researchers then manually analyzed the images to remove non-osteoclasts and add manual masks for osteoclasts that were missed. This process yielded 150 well images with ${\sim} 2\times 10^5$ mouse osteoclast masks and 24 slide images with ${\sim} 4\times 10^4$ mouse human osteoclasts masks. 

Mouse osteoclast images were split into five batches, M1, M2, M3, M4, and M5; and the human osteoclast images into two batches, H1 and H2; based on the experimental procedures, each consisting of tens of images. See \cref{fig:dataset} for samples. Since each batch potentially shares qualitative characteristics, we used these divisions for training and testing splits (\textit{e.g.}, we report mouse results under 5-fold cross-validation in \cref{sec:results}) and in our published dataset. However, since the scientific assays are ongoing, we did not indicate which batch represents which experimental procedure, and we did not conduct our computational experiments or analyses based on those distinctions.

\subsection{Osteoclast-Specific Nuclei Dataset}
\label{sec:nuclei_dataset}
We experimented with various nuclei instance segmentation datasets \cite{mahbod2024nuinsseg, kumar2017dataset, graham2021lizard, sirinukunwattana2016locality, wienert2012detection} in our osteoclast instance segmentation pretraining paradigm, but found them to be ineffective due to several notable differences between the biology and imaging of those nuclei and the nuclei in our osteoclast image datasets. Existing nuclei datasets use hematoxylin and eosin (H$\&$E) staining to improve the contrast of cell visibility under a microscope---it stains nuclei in a dark color, against a lighter cytoplasm, easing nuclei identification. By contrast, our TRAP-stained images show cells with light colored nuclei on a darker cell that is set against a light background. Consequently, nuclei detection methods developed for existing H$\&$E stained datasets did not perform well on our osteoclast images, prompting us to develop our own osteoclast-specific nuclei dataset.

Nuclei far outnumber osteoclasts and annotating each with segmentation masks would be infeasible for our purposes. Instead, we randomly selected 20 images from each of our five mouse experiments to curate a smaller nuclei dataset for manual annotations. In each image, we marked the approximate center-of-mass of each nucleus's shape with a point annotation tool VGG Image Annotator \cite{dutta2019via}. These annotations were converted to bounding boxes with width and height uniformly set to $36$ pixels, chosen empirically to cover larger nuclei in the dataset. A total of 20,784 nuclei were marked in this way, with an average of 200 annotations per image, illustrating the density of the information contained in the small dataset. By comparison, the well-known MoNuSeg dataset \cite{kumar2017dataset} has a similar count of 21,000 instances, albeit with full instance segmentation masks.

\section{Experiments \& Results}
\label{sec:results}

\begin{figure*}
    \centering
    \includegraphics[width=0.95\linewidth]{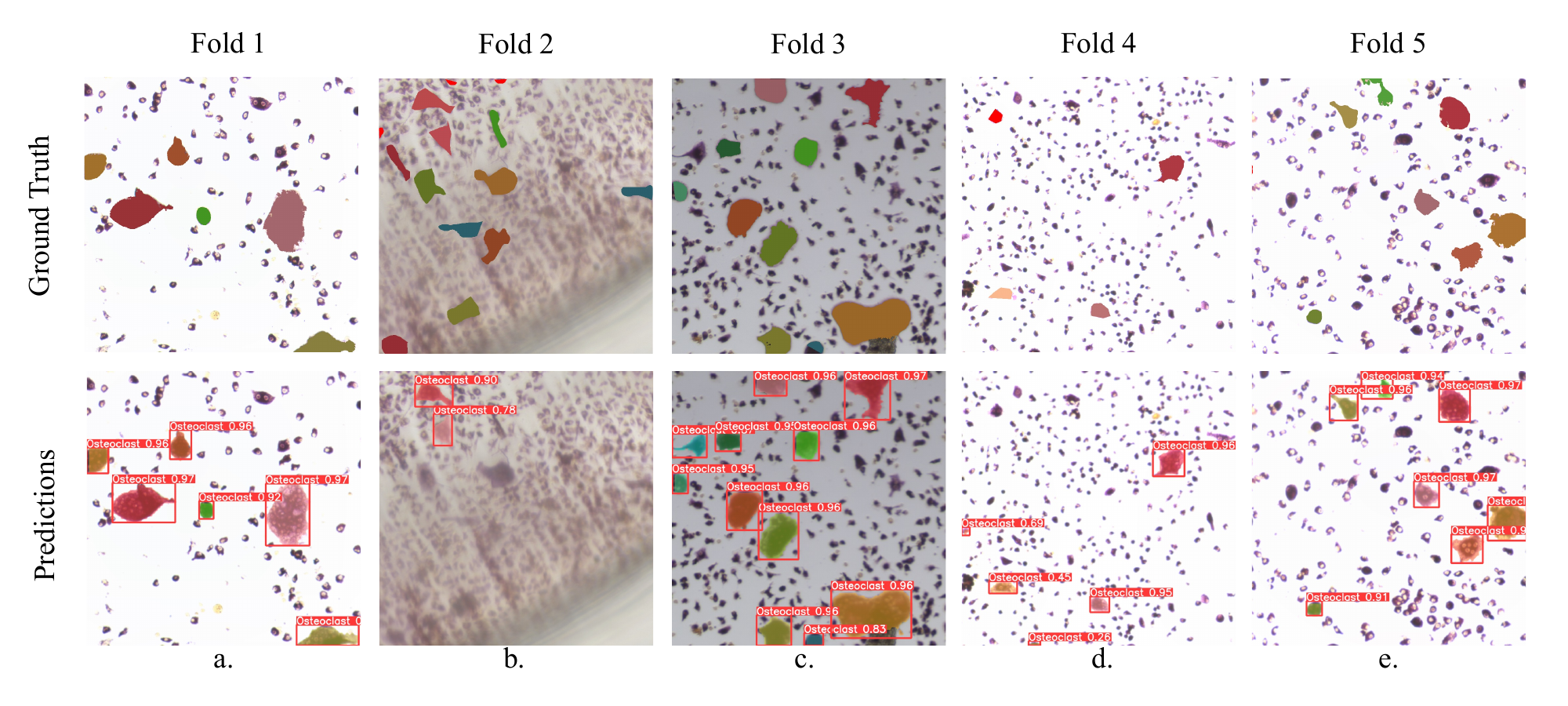}
    \vspace{-3mm}
    \caption{Predictions of our mouse osteoclast instance segmentation models under five-fold cross-validation. Ground truth osteoclast shape masks are carefully produced by experts. Our model's predictions of these masks are generally faithful, but we highlight some mistakes and challenges. In (b), image blur seems to impair the model's overall ability; in (c), a pre-osteoclast on the top left is mistakenly identified as an osteoclast; and in (d), an osteoclast on the top left is not detected, while two pre-osteoclasts below are mis-identified as osteoclasts.}
    \label{fig:fold_samples}
\end{figure*}

\begin{table*}
    \centering
    \caption{Mouse Osteoclast Instance Segmentation Model Cross-Validation (YOLOv8 M$\rightarrow$M)}
    \resizebox{0.9\linewidth}{!}{
    \ra{0.8} 
    \begin{tabular}{llrrrrrrrrrrrrrrrrrrr}
        \toprule
        \diagbox[width=25mm, height=5mm]& \textbf{IoU $t$} & \multicolumn{3}{c}{\textbf{0.10}} & \phantom{a} & \multicolumn{3}{c}{\textbf{0.25}} & \phantom{a} & \multicolumn{3}{c}{\textbf{0.50}} &   & \multicolumn{3}{c}{\textbf{0.75}} & & \multicolumn{3}{c}{\textbf{0.90}}\\
        \cmidrule[0.5pt]{3-5} \cmidrule[0.5pt]{7-9} \cmidrule[0.5pt]{11-13} \cmidrule[0.5pt]{15-17} \cmidrule[0.5pt]{19-21}
       \textbf{ Model} & \textbf{Test} & \textbf{P} & \textbf{R} & \textbf{mAP} & & \textbf{P} & \textbf{R} & \textbf{mAP} & & \textbf{P} & \textbf{R} & \textbf{mAP}  & & \textbf{P} & \textbf{R} & \textbf{mAP} & & \textbf{P }& \textbf{R} & \textbf{mAP}\\
         \midrule

         Fold 1 & M1 &  0.871&   0.838 & 0.906&  &  0.868      & 0.835& 0.902&  & 0.869 & 0.809 & 0.885  & & 0.816      & 0.714&0.784 & & 0.609      & 0.434&0.437\\
         Fold 2 & M5 &  0.884      &  0.524&  0.639& &  0.882      &  0.518& 0.630& & 0.851 & 0.487 & 0.583   & & 0.555      & 0.297&0.299 & & 0.063& 0.023&0.010\\
         Fold 3 & M4 &  0.894       &   0.850& 0.911&  &  0.892      &  0.845& 0.906& & 0.884 & 0.818 & 0.881   & & 0.831      & 0.722&0.779 & & 0.569      & 0.436&0.407\\
         Fold 4 & M3 &  0.862      &   0.826& 0.888&  &  0.858       &  0.820& 0.882& & 0.851 & 0.787 & 0.851   & & 0.805      & 0.677&0.735 & & 0.537      & 0.392&0.353\\
         Fold 5 & M2 &  0.908      &   0.849& 0.921&  &  0.907      &  0.845& 0.917& & 0.896 & 0.825 & 0.898  & & 0.852      & 0.736&0.812 & & 0.654      & 0.494&0.512\\
         \midrule

          \multicolumn{2}{l}{Average} & 0.884& 0.777& \textbf{0.853} && 0.881 & 0.773 & \textbf{0.847} && 0.870 &0.745 & \textbf{0.820}	&&0.772 & 0.629 &\textbf{ 0.682} && 0.486 & 0.356 &\textbf{0.344}\\
          \multicolumn{2}{l}{Standard Deviation} & 0.018& 0.142&  0.120&&  0.019&  0.143&  0.122&&  0.020& 0.145&	0.133&& 0.122&  0.187&  0.216&&  0.241&  0.190&0.195\\

         \bottomrule
    \end{tabular}
    }
    \label{tab:crossval}
\end{table*}

\subsection{Experimental Implementation}
Throughout, we used the large variant in the YOLOv8 family of models, YOLOv8l-seg, which has 45.9 M parameters in the backbone, detection, segmentation heads collectively. Each model was trained for 100 epochs, stopping early when validation split performance did not improve. Since the whole slide images are of gigapixel resolutions, following \cite{kohtala_automated_2022}, we extracted smaller patches of size $832\times832$ with 50\% overlap for our experiments. Images were downsampled to $416\times416$ for training and inference during pre-processing to speed-up inference, and outputs were scaled back to full resolution in post-processing. Other parameters were set to their default values \cite{Jocher_Ultralytics_YOLO_2023}. Performance was evaluated using standard precision, recall, and mean average precision ($\text{mAP}_t$ at various thresholds $t$) metrics from the Common Objects in Context (COCO) object detection dataset \cite{COCO_eval,COCO_API}.

We recall from our discussion in \cref{sec:background} that direct performance comparisons with earlier osteoclast image algorithms are difficult because groups addressing standard tasks like detection and semantic segmentation have not published their data and code. We hope our work can establish a public framework and benchmark for future machine learning work on osteoclast image analysis.

\subsection{Mouse Osteoclast Instance Segmentation}

\begin{figure*}
    \centering
    \includegraphics[width=0.95\linewidth]{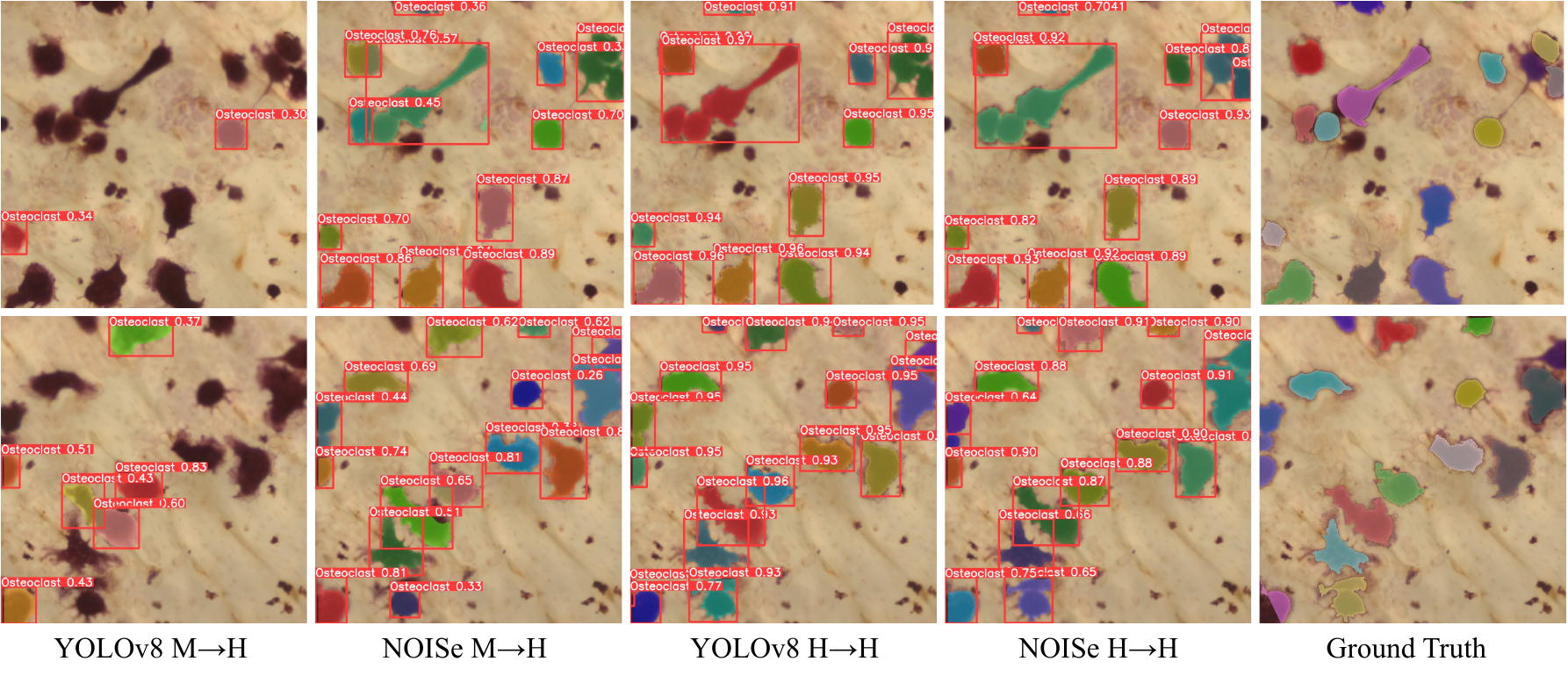}
    \vspace{-2mm}
    \caption{Qualitative results from different training configurations of YOLOv8 and NOISe models tested on human osteoclast data, showing that the NOISe strategy can improve performance on human osteoclasts even when the model is trained only on mouse data (NOISe M$\rightarrow$H).}
    \vspace{-2mm}
    \label{fig:noise_qual}
\end{figure*}

\begin{table*}
    \centering
    \caption{Human Osteoclast Instance Segmentation Performance under Various Train--Test Configurations}
    \resizebox{0.9\linewidth}{!}{
    \ra{0.8} 
    \begin{tabular}{llrrrrrrrrrrrrrrrrrrr}
        \toprule
        \textbf{IoU $t$} & \phantom{a} & \multicolumn{3}{c}{\textbf{0.10}} & \phantom{a} & \multicolumn{3}{c}{\textbf{0.25}} & \phantom{a} & \multicolumn{3}{c}{\textbf{0.50}} & \phantom{a} & \multicolumn{3}{c}{\textbf{0.75}} & \phantom{a} & \multicolumn{3}{c}{\textbf{0.90}}\\
        \cmidrule[0.5pt]{3-5} \cmidrule[0.5pt]{7-9} \cmidrule[0.5pt]{11-13} \cmidrule[0.5pt]{15-17} \cmidrule[0.5pt]{19-21}
         \textbf{Configuration} & \phantom{a} & \textbf{P} & \textbf{R} & \textbf{mAP} & & \textbf{P} & \textbf{R} & \textbf{mAP} & & \textbf{P} & \textbf{R} & \textbf{mAP}  & & \textbf{P} & \textbf{R} & \textbf{mAP} & & \textbf{P }& \textbf{R} & \textbf{mAP}\\
        \midrule
        YOLOv8 M$\rightarrow$H & &0.839 & 0.605 & 0.777 & & 0.826 & 0.595 & 0.755 &  & 0.695 & 0.666 & 0.595 &  & 0.425 & 0.309 & 0.285 &  &  0.084 & 0.098 & 0.046\\
        NOISe M$\rightarrow$H & &0.868 & 0.783 & \textbf{0.878} & & 0.861 & 0.775 & \textbf{0.865} &  & 0.812 & 0.729 & \textbf{0.796} &  & 0.677 & 0.586 & \textbf{0.592} &  &  0.324 & 0.233 & \textbf{0.151}\\ \midrule
        YOLOv8 H$\rightarrow$H & &0.849 & 0.826 & 0.884 & & 0.843 & 0.816 & 0.872 &  & 0.799 & 0.766 & 0.805 &  & 0.702 & 0.609 & 0.626 &  &  0.506 & 0.339 & \textbf{0.296}\\
        NOISe H$\rightarrow$H & &0.860 & 0.810 & \textbf{0.891} & & 0.857 & 0.800 & \textbf{0.880} &  & 0.807 & 0.757 & \textbf{0.820} &  & 0.722 & 0.613 & \textbf{0.650} &  &  0.478 & 0.320 & 0.265\\
        \bottomrule
    \end{tabular}
    }
    \label{tab:human_transfer}
\end{table*}

We tested the model on the mouse datasets via five-fold cross-validation. For each fold, one dataset was set aside for testing, and of the remaining training data, a random 20\% was reserved for validation. Cross-validation result are reported in \cref{tab:crossval}. At the default threshold, $t=0.5$, the model maintained an mAP score above 80\% on all test sets except M5. Performance is weaker under the most severe threshold, $t=0.9$, but is strong overall. Illustrated examples of predictions on each cross-validation fold can be found in \cref{fig:fold_samples}. In general, the model does an excellent job of predicting the masks of identified objects, but it can sometimes struggle to distinguish between osteoclasts and pre-osteoclasts (which are similar to osteoclasts but have fewer than three nuclei), or to detect osteoclasts when the image is blurry. The weak performance on the M5 test set in particular could be caused by blurring artifacts present in some images, such as the one shown in \cref{fig:fold_samples}.

\subsection{Transfer Learning for Human Osteoclasts}
\label{sec:human-results}

In \cref{tab:human_transfer}, we compare the instance segmentation performance of our NOISe model against the baseline YOLOv8 model, under various mouse--human and train--test configurations to quantify both models' generalizability. This also serves as an ablation study for the NOISe method. The configurations are indicated by a MODEL TRAIN$\rightarrow$TEST naming scheme which we expand on below. 

In our notation, the baseline configuration employed in \cref{tab:crossval} would be denoted YOLOv8 M$\rightarrow$M, with the mouse-to-mouse domain indicator M$\rightarrow$M implying a cross-validation paradigm. The YOLOv8 M$\rightarrow$H configuration denotes a baseline YOLOv8 model trained on the entire mouse data and tested on all human data. The performance drop in YOLOv8 M$\rightarrow$H, compared to YOLOv8 M$\rightarrow$M, shows that YOLOv8 trained on mouse data does not generalize well to the human domain. By contrast, performance is much stronger under the NOISe M$\rightarrow$H configuration, where YOLOv8 is trained with all of the mouse data under the NOISe strategy and tested on all of the human data. This demonstrates effectiveness of NOISe in improving generalization to unseen human data.

The bottom half of \cref{tab:human_transfer} shows results from both YOLOv8 and the NOISe model under the H$\rightarrow$H configuration of two-fold cross-validation on the human data (with performance on both folds averaged). Contrasting the top and bottom halves of \cref{tab:human_transfer} shows that training on some human data improves test performance on human data, as expected. But more interestingly, focusing on the bottom half of \cref{tab:human_transfer}, we see that when some human data is available for training, the NOISe strategy can still offer further, small gains---yielding the strongest overall performance on human models---but these gains are nowhere as dramatic as those provided by NOISe in M$\rightarrow$H setting. 

We show some qualitative results in \cref{fig:noise_qual}. These illustrate the improved osteoclast detection ability (recall) of NOISe M$\rightarrow$H over YOLOv8 M$\rightarrow$H, and closer inspection shows that when predicting correct instances that both agree on, NOISe does so with higher confidence. The YOLOv8 H$\rightarrow$H and NOISe H$\rightarrow$H predictions are more comparable to each other. They are also comparable to those of NOISe M$\rightarrow$H, which is again remarkable, since the latter has never seen any human osteoclast data (nor human nuclei data). These numerical and qualitative results show that the NOISe strategy significantly improves mouse-to-human domain transfer, and suggests that it might generalize effectively to other, new osteoclast image domains. Thus, we recommend NOISe M$\rightarrow$H and NOISe H$\rightarrow$H as starting points in such settings.

\section{Conclusion}
\label{sec:conclusion}
We have presented the first osteoclast instance segmentation dataset and algorithm, fully automating the laborious cellular image analysis task performed by osteoporosis researchers. We introduced a novel nuclei-aware pretraining strategy which significantly improves the performance of models trained only on mouse osteoclast data, when tested on human osteoclast images, which are harder to obtain and annotate. To support further algorithm development and accelerate osteoporosis research, we publish our annotated mouse osteoclast dataset and the code and pretrained weights for our NOISe model, which appears to exhibit strong generalizability and is recommended for both familiar and new osteoclast image settings. Future work could explore gains in performance and generalizability with more robust nuclei supervision, or by incorporating methods like cell stain style transfer to address variability in lab preparation or image appearance. Continual ``human-in-the-loop'' learning could enable more efficient gains in segmentation performance, and improve domain transferability on a larger scale.

{
    \small
    \bibliographystyle{ieeenat_fullname}
    \bibliography{main}
}


\end{document}